\documentclass[final, twocolumn]{article}

\usepackage{mathptmx}
\usepackage{graphicx}
\usepackage{subfig}
\usepackage{url}
\usepackage{listings}

\title{Prediction of Wort Density with LSTM Network}
\author{Derk Rembold, Bernd Stauss, Stefan Schwarzkopf}

\begin{document}

\maketitle

\begin{abstract}
	Many physical target values in technical processes are error-prone, cumbersome, or expensive to measure automatically. One example of a phyhical target value is the wort density, which is an important value needed for beer production. 
This article introduces a system that helps the brewer measure wort density through sensors in order to reduce errors in manual data collection.
Instead of a direct measurement of wort density, a method is developed that calculates the density from measured values acquired by inexpensive standard sensors such as pressure or temperature. The model behind the calculation is a neural network, know as LSTM.

\end{abstract}


\section{Introduction}\label{introduction}

Many manufacturing processes' physical values require compliance with certain parameter in order to ensure that the product specification is met and that the product quality remains on a constant level.
This requires that these values can be reliably measured, which often means that expensive sensors need to be used. Measuring physical values during production can be complex to operate, install and maintain.

An example is the determination of the wort density in the enzymatic conversion of starch in a fermentation process, as occurs in beer production.

When determining the wort density in liquids, the refractometer or, preferably, the spindle (saccharometer) is usually used. The spindle itself shows directly the wort density, and indirectly the sugar content.
These two measuring processes cannot be carried out "inline", i.e. during the ongoing production process. They require samples to be taken regularly, for which laboratory staff manually determines the actual measured variable.

This procedure is rather unsuitable for continuous measurement with the aim of keeping an eye on the measured variable throughout the entire process execution. Furthermore, it is very error-prone e.g. due to incorrect metering or transmission to a recording system. Lui et. al. \cite{bib24} point out that this phenomenon is widely spread in practice and propose a so called soft sensor modeling that can help to overcome those difficulties. Indeed, soft sensors have proven to be a pragmatic approach: Instead of a direct measurement of values of interest, other values that are physically easier to determine with less expensive sensors are captured and the target variable is derived from these.

\section{Basics}\label{basics}

\subsection{Data Gathering}\label{datagathering}

When designing the experimental setup for the brewing process, two major ideas stood in the foreground. First, experimental design should be transparent to the laboratory staff at any time to prevent mistakes in the execution of the experiment.  
Second, the experiment should be operated in a reliable way. This means, routine tasks, which can not run automatically, will be assisted by the system.   

\begin{figure}[h]%
	\centering
	\includegraphics[width=0.5\textwidth]{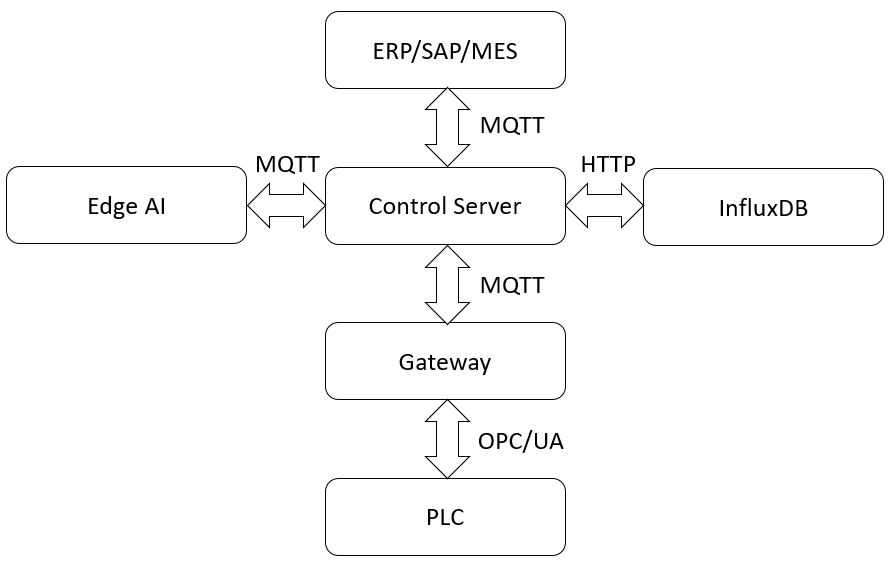}
	\caption{Architecture}\label{sn-arch-1}
\end{figure}

To describe the setup in more detail, we will refer to Figure~\ref{sn-arch-1} which shows the technical design at the lowest level.
For the brewing process, the researchers use equipment of the company Speidel that provides home brewery devices for private customers. Although this ready-to-use equipment comes already with nearly fully automated operation modes, its control unit was replaced due to the comprehensive, very specific requirements mentioned above.

Therefore, the technical process is controlled by an own 
programmable logic controller (PLC). Some sensor values, e.g. 
pressure values, power consumption of the actuators as well as 
the current temperature, can be captured via the PLC.

The various input data is collected in a gateway using different protocols and technologies, e.g. OPC-UA, RS232 or REST interface, and propagated to the central control server which take the role of a central data hub. Beside that, the gateway serves as a firewall as it decouples the brewing equipment from the external network.

The communication with the external components of the overall system is shown in Figure~\ref{sn-arch-1} . Communication takes place via the MQTT-protocol that is a standard protocol in many IOT applications. 
A MQTT broker, which is not explicitly shown in Figure~\ref{sn-arch-1}, receives messages from the connected components to publish their information for further access to its subscribers, such as the control server. The control server processes the messages, monitors the communication between the components and saves the sensor data in the InfluxDB database. 

The control server is the most central component to bridge the gap between the machine level on one side and the operational information system, i.e. SAP and the integrated manufacturing execution system (MES) on the other. 
The edge AI device has access to the sensor values and state
variables of the technical process via the MQTT-messages and the InfluxDB. This data is used for training purposes of the implemented artificial intelligence (AI). 

The system offers at the top level a graphical user interface (GUI), namely a system that enables production data acquisition via the integrated MES. The application allows them to define the process and interact with the other components. All the process parameters like ingredients, specifications, operations lists etc. are kept in the SAP ERP-system and are provided to the other components by MQTT-messaging. With the ERP-system, all logistic operations are made transparent in the overall system. Specifically, this includes the batches of the ingredients and their characteristics that were actually used in the brewing process and that usually vary from batch to batch although the material specification is clear.
In fact, this information is very critical since it is supposed to influence the prediction of the wort densitiy. 

With regard to the fermentation process of the wort, the same architecture, see Figure~\ref{sn-arch-1}, is used. However, the control of the process was much easier since no active intervention in the fermentation process was necessary.

\subsection{Brewing Equipment}\label{equipmentbrew}

\begin{figure}[h]%
\centering
\includegraphics[width=0.4\textwidth]{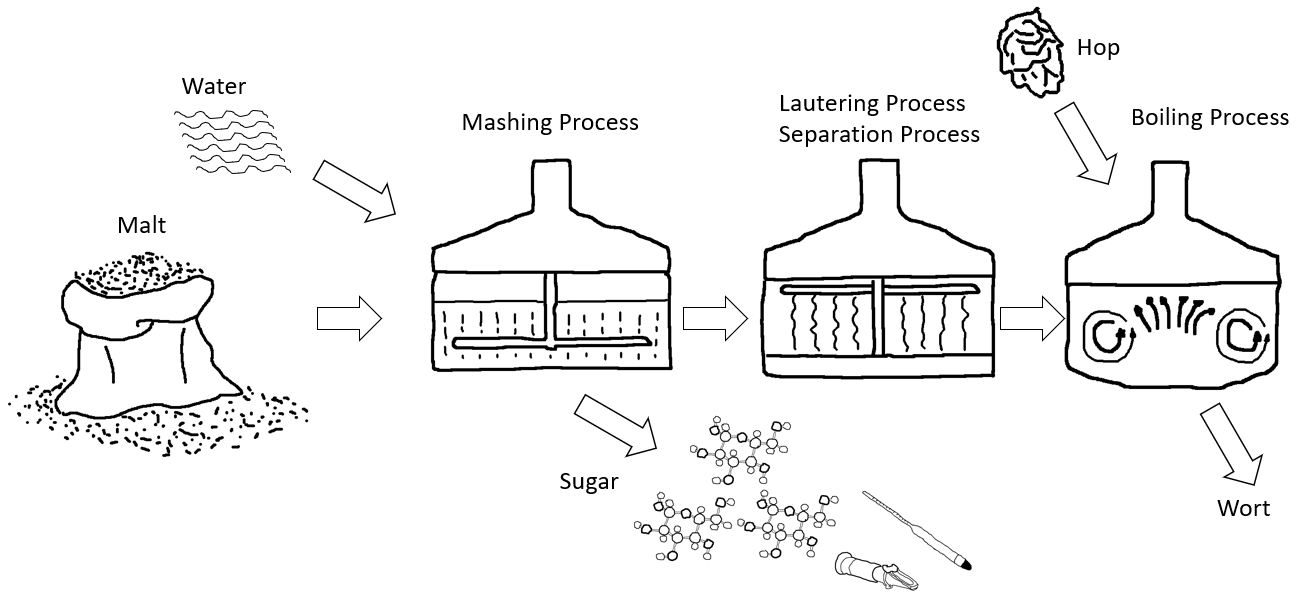}
\caption{Brewing Process}\label{sn-beb-1}
\end{figure}

In the following, the beer brewing process is described; more details can be found in \cite{bib16}, Chapter 7. The brewer fills in water and malt into the mash container, see Figure~\ref{sn-beb-1}. The brewer heats up the mash container to 61-65 degrees Celsius for about 20 minutes. During this process step, beta-amylase enzymes inside the malt convert its starch into sugar. The starch looses here its crystalline structure and cleaves to maltose. This biologic process is called gelatinization, the process step is called maltose rest.  Then the container is heated up to around 72 degree Celsius for about 10-20 minutes. At this temperature, the beta-amylase enzymes are unstable and get deactivated. Since the alpha-amylase enzymes are still stable at this temperature, they come into play and produce non-fermentable sugars (e.g. dextrins). The brewer calls both of these process steps saccharification rest. After the saccharification rest, the mash inside the container is heated up to 78 degrees Celsius to stop the enzyme activities and to destroy the remaining proteins. The rest time periods and temperatures mentioned above depend on the beer's recipe and on the brewer's knowledge. 
After the mashing process, the mash is filled into a lautering container to wash out the remaining sugar from the grains. Here, the brewer's equipment pours water heated to 78 degrees Celsius over the mash. To separate the mashes liquid from the solids, the brewer often uses a whirlpool. This equipment stirs the mash in a rotary motion, so a swirl effect will come up. The wort (the liquid part of the mash) accumulates at the side of the container and the solids of the mash in the middle of the container due to the centrifugal force. During this process, the purified wort is pumped into the cooking container. Finally, the temperature of the wort in the cooking container is brought up to boiling temperature and hop is added. The boiling length is determined by the time the hop ingredients (such as the alpha acids) need to dilute into the wort. After the boiling process, the brewer cools down the wort to below 20 degrees Celsius for the upcoming fermentation process.

\begin{figure}[h]%
\centering
\includegraphics[width=0.3\textwidth]{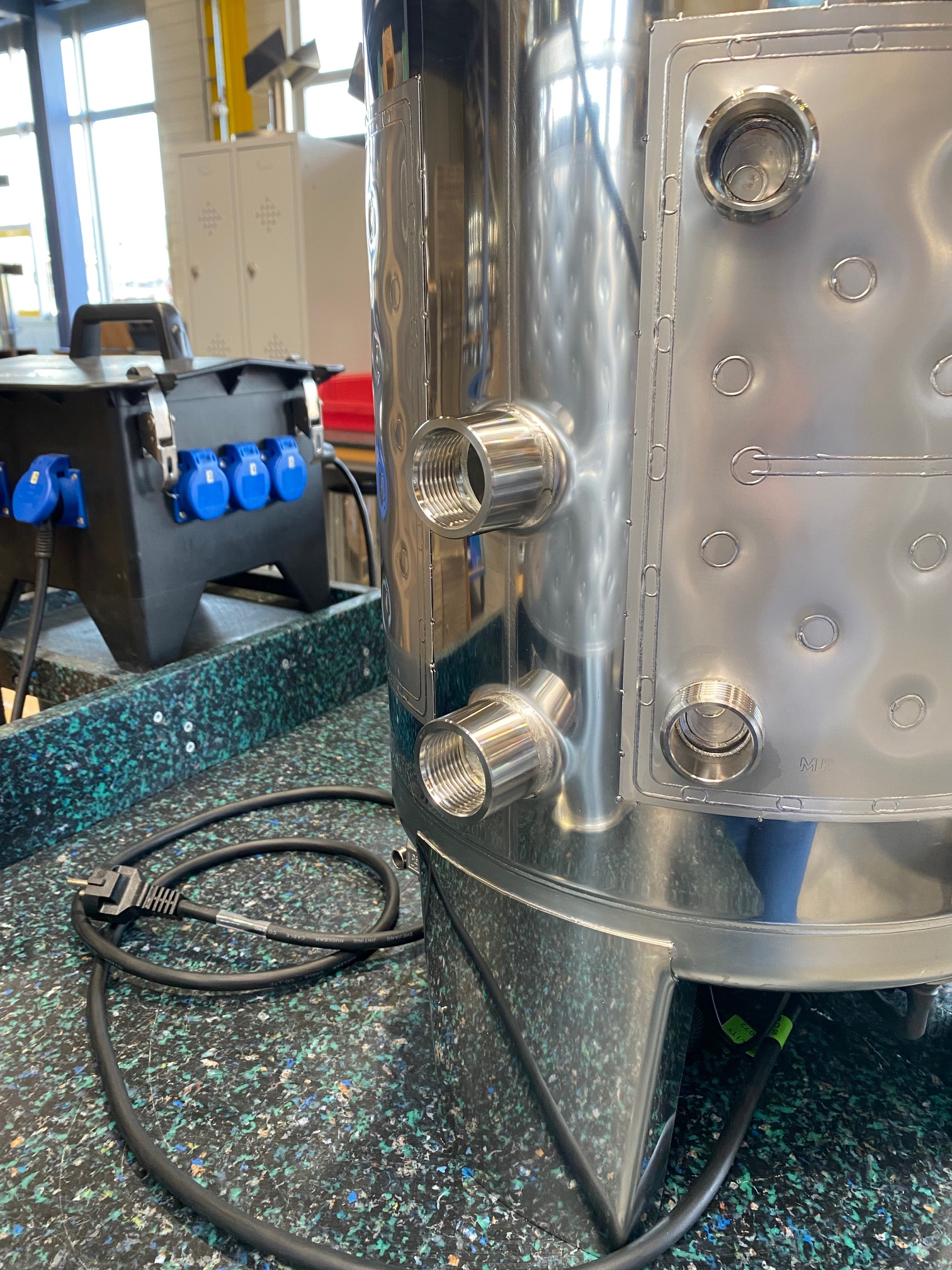}
\caption{Brew Container}\label{sn-beb-2}
\end{figure}

The researcher's equipment, that was a donation of the company Speidel, a producer of beer brewing equipment, can be used for all steps in the entire brewing process, i.e. mashing, lautering, purifying and cooking, see Figure~\ref{sn-beb-2}. 
The container has an attachment for a temperature sensor to measure the mash and wort temperature.

Since sugar concentration has an influence on the fluid density as well as the wort density, two pressure sensors were added. Figure~\ref{sn-beb-2} shows a beer brewing container with two sockets, welded at the side of the container, for two pressure sensors. The sockets height positions have a difference of 10 cm. With two pressure sensors and their height difference, together with the temperature sensors the researchers expected to deduct the wort density by calculating the fluid pressure (hydrostatic pressure).

\subsection{Fermentation Equipment}\label{equipmentferment}

\begin{figure}[h]%
\centering
\includegraphics[width=0.45\textwidth]{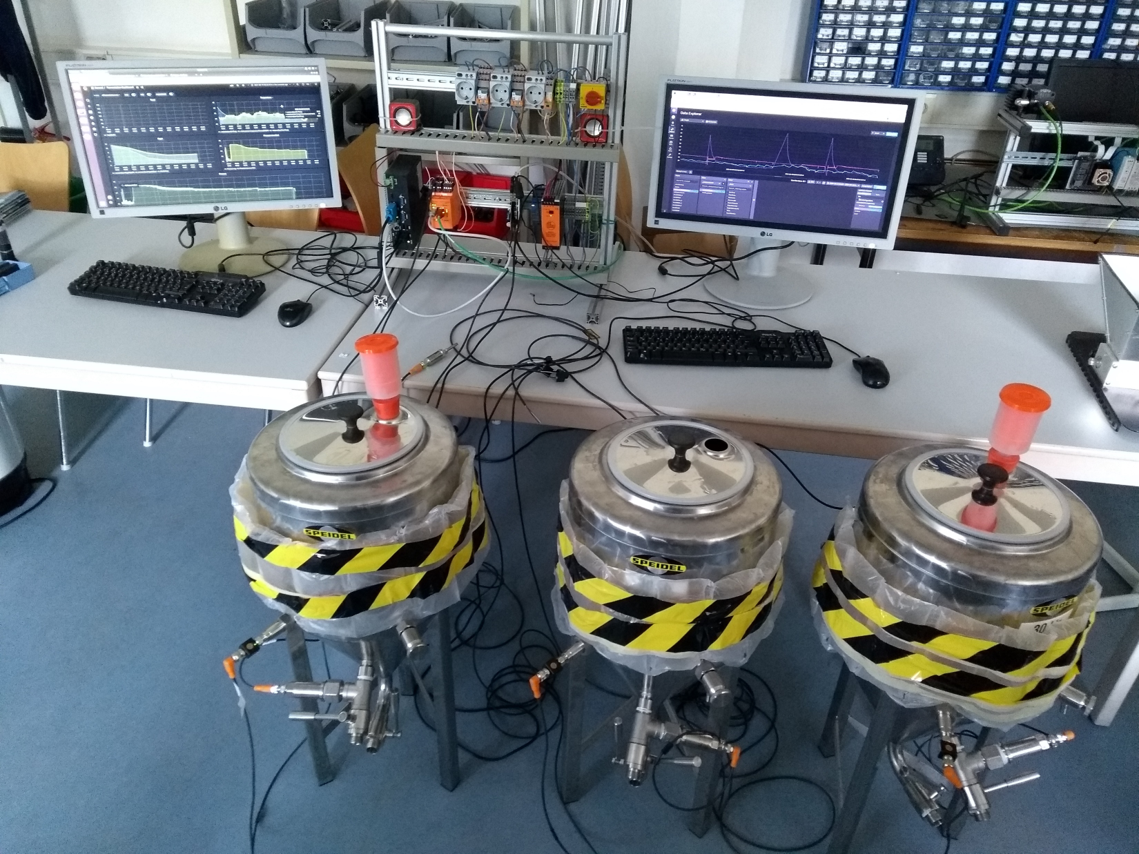}
\caption{Fermentation Containers}\label{bae-fig1}
\end{figure}

The researcher's experimental setup for the femernation process consists of three fermentation containers, which can be operated separately, see Figure~\ref{bae-fig1}. This allows to have three fermentation processes running in parallel. Since a fermentation process can take up to a week, the running processes will not hold up each other when new wort is produced. 

Each container has a pressure sensor at the bottom. A temperature sensor attached at the side of the container measures the wort temperature. All sensors have an IO-Link interface and are connected to the control server to gather the data into a time series database.

Each container has an airlock to prevent bacterial infections of the wort. Airlocks are filled up with water, which separates the container's atmosphere from the environment atmosphere. During the fermentation process, the yeast produces carbon dioxide which builds up a pressure and the airlock is lifted. If the pressure inside the container exceeds a threshold pressure, determined by the airlock's water height, carbon dioxide is released. Therefore, the pressure maximum inside the container is directly related to the height of the water filled into the airlock due to the hydrostatic pressure. The researcher filled exactly 5 cm of water into the airlock. The container's atmosphere has therefor maximum pressure of 50 mBar. The pressure sensor at the bottom of the container measures not only the container's atmosphere, but also wort's hydrostatic pressure from the wort. Both pressures add up at the sensor's output.  

Besides the pressure sensors, there is a temperature sensor to measure the environment temperature and a pressure sensor to measure  the air pressure of the environment. The control server also collects this sensor data and stores them to the InfluxDB. The InfluxDB is particularly suited for time-series data. Additionally, a grafana visualization platform is installed on the control server for visualizing the measured data.

\begin{figure}[h]%
\centering
\includegraphics[width=0.5\textwidth]{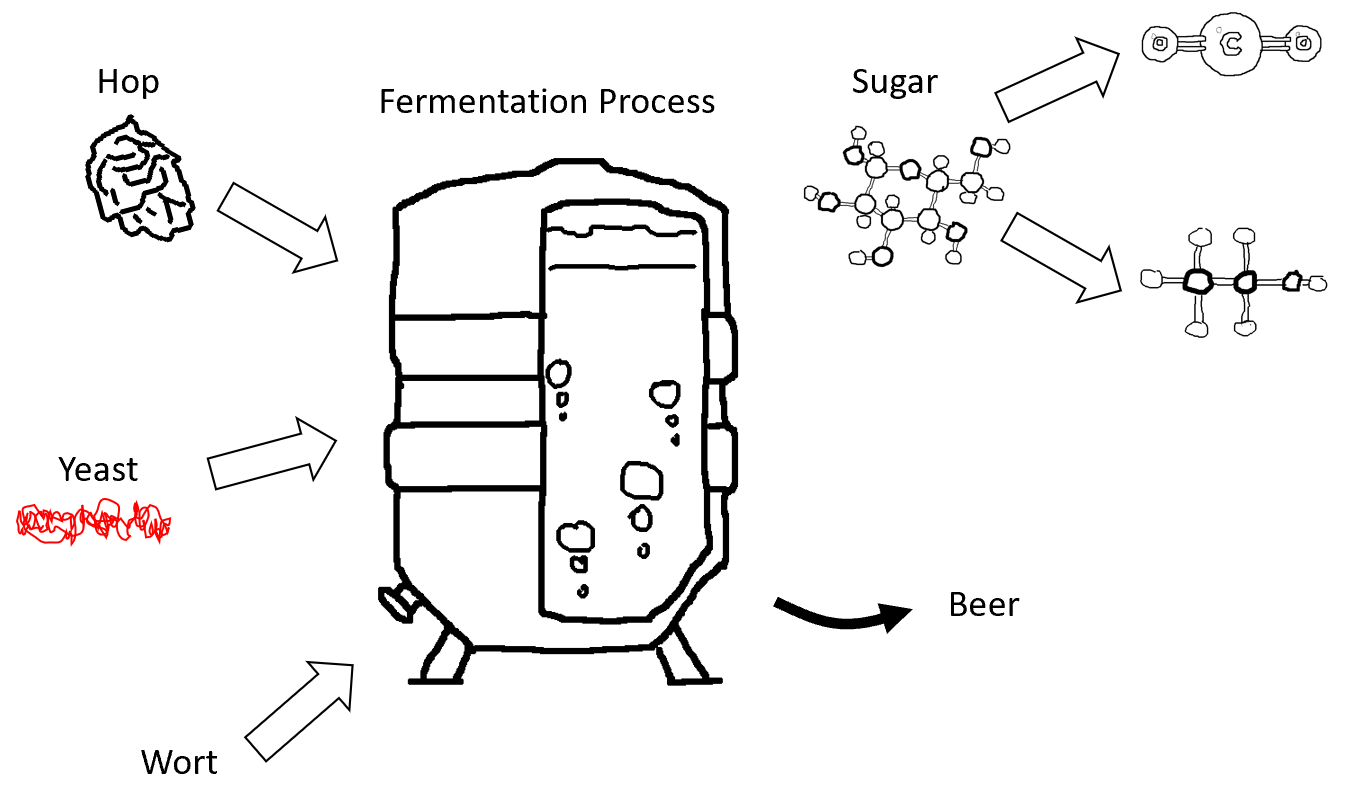}
\caption{Fermentation Process}\label{bae-fig2}
\end{figure}

In \cite{bib16}, Chapter 8 the author described the fermentation process in detail. However, for the following consideration a brief scatch of the single process steps suffices:
After the boiling process, wort is filled into one of the fermentation containers and it is cooled down to 20-30 degree Celsius. At this temperature, yeast has a good surviving condition. Yeast is added into the container and the fermentation process is starting, see also Figure~\ref{bae-fig2}. It is noted that the time between filling wort into a fermentation container and adding yeast must be as short as possible to avoid bacterial infections. The added yeast will consume the fermentable sugar and oxygen, and the yeast will multiply. After the oxygen inside the fermentation container is consumed, the yeast will switch into a survival mode. It continues to consume the fermentable sugar, but it also produces alcohol and carbon dioxide, which increases the pressure inside the container. The airlock releases carbon dioxide if a pressure threshold (50mBar) inside the fermentation container has been reached.

The brewer controls the wort density during the fermentation process with tools called spindle and refractometer. Both tools need training and additional equipment, such as probe cooling or probe heating devices, since the spindle and the refractometer are calibrated for specific probe temperatures. As pointed out in \cite{bib23} measuring correct data is cumbersome and error-prone (it needs training). The measured data is not online available, unless an industrial sensor is used, which is often very expensive to buy and to install. Also, the article \cite{bib23} proposes a trained neural network to determine the physical data which are difficult to measure.

\section{Prediction of wort density}\label{predictionferment}

\subsection{Prediction of wort density during the fermentation process}\label{predictionfermentprocess}

\begin{figure}[h]%
\centering
\includegraphics[width=0.5\textwidth]{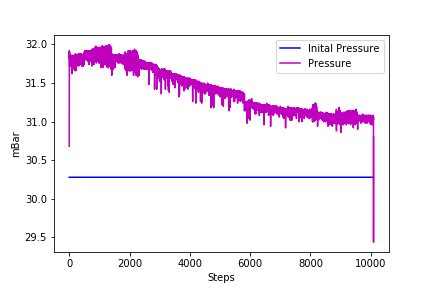}
\caption{Pressure during the fermentation process}\label{sn-posf-2}
\end{figure}

The fermentation process is a time-varying process, which means that process behavior is changing with time \cite{bib19}, \cite{bib26}, \cite{bib27}. This is especially observable at the beginning of the fermentation process, where the researchers observe in most experiments a drop of pressure within the first day after yeast was added. After this, a build-up of pressure after can be seen, as well. There are however some experiments, where the researchers noticed a constant level of pressure until the yeast really starts to produce carbon dioxide, see Figure~\ref{sn-posf-2} before Steps 1000.
After this pressure drop and pressure build-up phase, the amount of wort and the sugar content is decreasing steadily since the yeast starts its work to produce alcohol and carbon dioxide. The carbon dioxide is released through the airlock and therefore the pressure sensors measure a decreasing fluid pressure (hydrostatic pressure), see Figure~\ref{sn-posf-2}. This process will continue until the yeast has completed to eat the fermentable sugar inside the wort. This process takes about 5 to 7 days. Due to the time variance, it was decided to capture the time information inside a variable. First, the time variable is set to 0 and every 60 seconds the control server takes a new measurement from all sensors and the time variable is increased by one. In Figure~\ref{sn-posf-2} the reader sees this time variable called $Steps$ in the x-axis.

\begin{figure}[h]%
\centering
\includegraphics[width=0.5\textwidth]{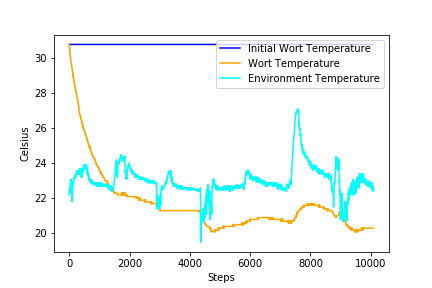}
\caption{Temperatures during Fermentation}\label{sn-posf-3}
\end{figure}

Other measurement data besides the pressure of the fermentation container is the wort temperature and the environment temperature, see Figure~\ref{sn-posf-3}. The reader can see that the wort temperature is still cooling down after wort from the boiling process is pumped into the fermentation container. Figure~\ref{sn-posf-3} shows after $Step$ 2000 fluctuations of the wort temperature. The fluctuations are due to the environment temperature change during day and night, which also heats up and cools down the content of the fermentation container.  

Spindles and refractometers have scale values between 0 and 20 with units called Plato or Brix. The higher the value, the higher the density and therefore the higher the content of fermentable and non-fermentable sugar and other components besides water.

Since the researcher's goal was to determine the density (and indirectly the sugar content) with pressure and temperature sensors to omit the cumbersome handling with spindles or refractometers a prediction function is required that calculates the wort density from the captured sensor data. A quite generic approach that is used here takes a neural network, that however must be trained. 

\begin{figure}[h]%
\centering
\includegraphics[width=0.5\textwidth]{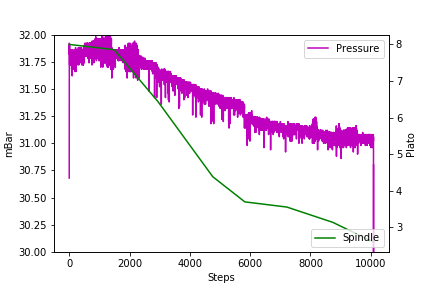}
\caption{Pressure and Sugar Content}\label{sn-pic-Pressure-Spindle}
\end{figure}

Therefore, in addition to the sensor data, target values had to be collected by measuring the wort density using spindles and refractometers. For the refractometer measurement, only a small probe is needed. However, the spindle needs a 100ml probe to get valid results. Unfortunately, removing a 100ml probe from the container will show an effect on the pressure sensor. The pressure curve in Figure~\ref{sn-pic-Pressure-Spindle} shows a drop of pressure  at around $Step$ 5500. This is the time, when the lab assistant took a spindle probe. The drop in pressure during wort density measurement is the reason why one more variable was introduced, i.e. an event-information, that tracks when a spindle probe was taken.

\begin{figure}[h]%
\centering
\includegraphics[width=0.5\textwidth]{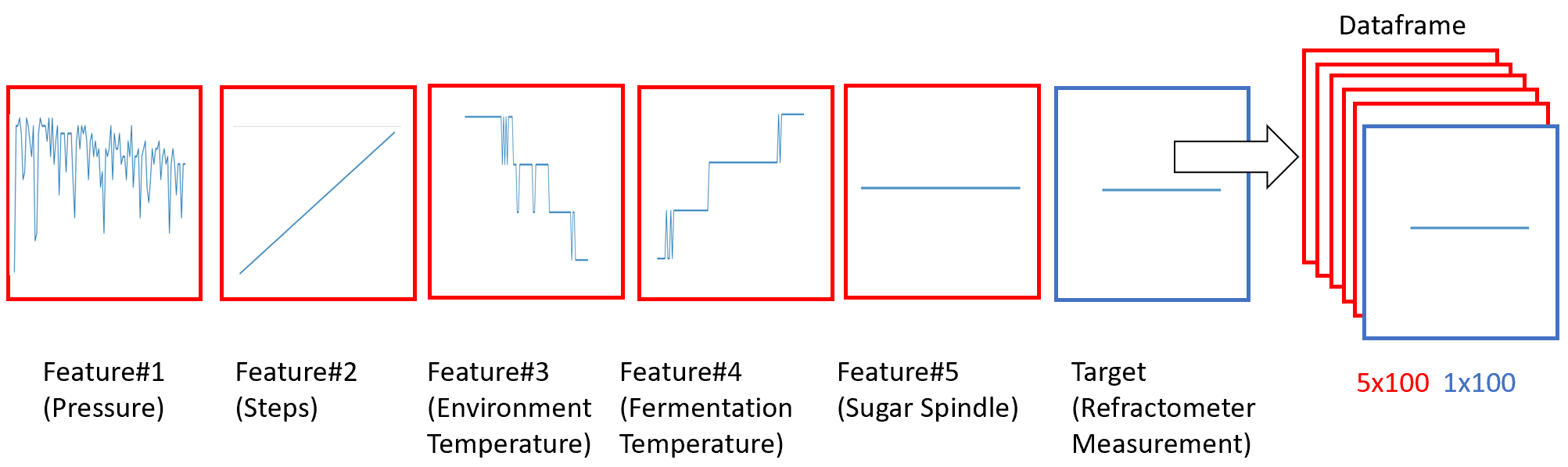}
\caption{Dataframe}\label{sn-posf-1}
\end{figure}

Figure~\ref{sn-posf-1} depicts how the data are stacked to a data frame. The reader can see there are three sensor values: Feature\#1 (pressure), Feature\#3 (environment temperature), and Feature\#4 (fermentation temperature), the time information Feature\#2 ($Steps$) and the event information Feature\#5 ($Sugar Spindle$). The target value is the spindle and refractometer probe value measured in Plato, see Figure~\ref{sn-posf-1}. To sum up, in one data frame there are data values of five features and one target.

Each feature in a data frame has 100 sequential data points. Since the control server pulls every 60 seconds data points from all sensors, the length of time of a data frame is $100*60$ seconds. All sensor values are stacked to create the final data frame, which is a 100x6 matrix. One matrix forms a single training data set to be used as input for the neural network.

\cite{bib25} depicts similar input values for training a neural network, which are pH, CO2 concentration, fermentation temperature, time, initial concentration of substrate. The researchers have intentionally left out pH values, since there is no significant change in pH. The researchers will discuss the initial concentration of the substrate (initial density) in Chapter~\ref{conclusion}.

\begin{figure}[h]%
\centering
\includegraphics[width=0.5\textwidth]{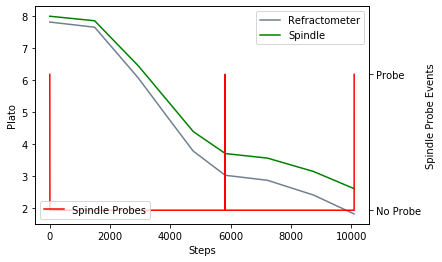}
\caption{Refractometer Measurements and Spindle Extrapolations for Target Data}\label{sn-posf-4}
\end{figure}

Above in this Chapter, the researchers have pointed out that spindle measurement affects the weight of the wort inside the fermentation container and therefore has an effect on the values taken by the pressure sensor. This is the reason not more than five spindle probes were taken during a single fermentation process. However, in each fermentation experiment the researchers vary the number of spindle probes. Since the refractometer measurement does not influence the wort's weight (only few drops of probe are taken), the researchers can take each day several refractometer probes. Unfortunately, the measurement with refractometers is less precise and more error-prone. This is the reason both methods are used, spindles and refractometer to gather the target values. The red curve (spindle probes) in Figure~\ref{sn-posf-4} shows three events where spindle measurement were taken (at Step 0, Step 5500 and Step 10000). The refractometer curve in gray shows the interpolated curve from the refractometer data (the unit is in general in Brix, but it can be easily concerted to Plato). Every day (or even more often) a refractometer probe was taken from the wort. The curve data between the probe data values were interpolated. Unfortunately, there are too few spindle values to draw a curve with such a precision like the refractometer curve, so the researchers extrapolated the spindle curve with the refractometer data (while taking the units Plato and Brix into account).  Figure~\ref{sn-posf-4} shows the spindle curve in green, which is actually a processed curve from spindle data and from refractometer data in combination.

At the beginning of Chapter~\ref{predictionferment}, it was explained how the feature data (sensor data, time information, spindle probe events) and target data (extrapolated spindle data) are stacked to a data frame with size 6x100 (5x100 features and 1x100 target). Each row in the data frame corresponds to all sensor values at a point in time, and there are 100 sequential points of data with a sample rate of 60 seconds. One data frame captures therefore $60*100$ seconds of feature and target data. The data of two sequential data frames are overlapping, which means that the values of a starting point in one data frame has already been used in a previous data frame. The distance in $Steps$ between two starting points of two data frames is captured in the variable $overlap$.

During the project, 31 charges of beer have been brewed and fermented in sum. Since each fermentation experiment takes about 5-7 days (in average 10000 $Steps$), this results in about 10000 data frames for just one fermentation process, if a maximum of overlapping ist used, i.e. $overlap = 1$. Unfortunately, there is a compromise one has to find here due to the limited number of experiments. 
In order to have as much training data as possible, a smaller $overlap$ would be desirable. However, since in these cases the training data of two adjacent data frames do not differ a lot, the neural network is thus trained to some extent with similar data. On the other side a large $overlap$ value will reduce the amount of training data while having more variation in the training data. 

The authors decided you have ca. 40000 data frames to have a sufficient amount of training data. Each time new data from a fermentation experiment is produced, the $overlap$ value was adapted. As mentioned above at the end of the project the researchers executed 31 experiments with around 10000 data captured each, so the researchers chose to have $overlap=7$ at the end of the project.

Having the data frames for training, the researchers needed to choose a model to predict the wort density value. In \cite{bib21} the article's author proposes a method to estimates biomass concentration during the fermentation process using a LSTM model. Here the author uses input parameters such as pH, glucose values, initial substrate concentration. In the article \cite{bib18} the authors predicted analyte concentrations of a fermentation process using statistical methods. They combined infrared spectroscopy data, chemical data and physical data. They claim that the root-mean-square error was improved by 50\%. In \cite{bib20} sensor data were fused using a neural network to predict and monitor variables such as lactose during the process of yogurt fermentation. Sensor input data had been pH, near-infrared spectroscopy spectral data (NIRS), electronic nose data, etc. It was shown that quality of process control and monitoring has been improved. 

The nature of the fermentation process is strongly dynamic.

Proof that this is a time variant process. See also above. The researchers show this in their figures, where the reader can see that the pressure inside the fermentation container is building up shortly after the yeast is added and continues to decrease while the yeast eats the sugar and produces alcohol and carbon dioxide.

In the previous work \cite{bib14}, the authors have shown that LSTM neural networks can give good results for this kind of experimental setups. Another article \cite{bib31} shows good results for the same kind of problem solution as well. In \cite{bib22} the authors propose a LSTM model to estimate oxygen content in gas produced in boilers from a coal-fired power plant. They show that their LSTM model creates good generalization results. The authors in \cite{bib26} argue that key biological variables are difficult to measure during fermentation processes and propose to use neural networks to overcome the problem. In this article, the fermentation process for penicillin production is described. The target value is the substrate and product concentration. 

An early work on modelling a fermentation process with neural networks and time delay connections have been published in \cite{bib27}. Similar to LSTMs, a neural network with delay connections inside the neural network were proposed.

Due to the successful use of LSTM Networks, the researchers decided to use this kind of network as well. Listing~\ref{posf-listing-Keras} shows Keras code to implement a LSTM neural network, described in the Keras instructions in \cite{bib15}.

\begin{minipage}{\hsize}%
\lstset{frame=single,framexleftmargin=-1pt,framexrightmargin=-17pt,framesep=12pt,linewidth=0.98\textwidth,language=python}
\begin{lstlisting} [caption={Keras Code},captionpos=b, label={posf-listing-Keras}]
inp = Input(shape=
	(windowsize, totalfeatures))
x = LSTM(dim)(inp)
x = Dense(128, activation='relu')(x)
x = Dense(64, activation='relu')(x)
outp = Dense(1, 
	activation = 'linear')(x)
model = Model(inputs=inp, 
	outputs=outp)
\end{lstlisting}
\end{minipage}

The neural network has an input layer with a matrix as input, described by variables $windowssize$ and $totalfeatures$. Note, a data frame from this project's experiments have the size 100x6. Since one column is the target, the researcher set $totalfeatures$ to 5 and $windowssize$ to 100. The variable $dim$ describes the number of LSTM elements, which the researchers have set to four. Two more fully connected layers follow. The last layer (and output layer) has one neuron with a linear activation function. The neural network outputs one value describing the wort density (and indirectly the sugar content) from the data frame.

\subsection{Results}\label{results}

\begin{figure}[h]  
  \begin{minipage}[b]{0.5\linewidth}
    \includegraphics[width=0.85\textwidth]{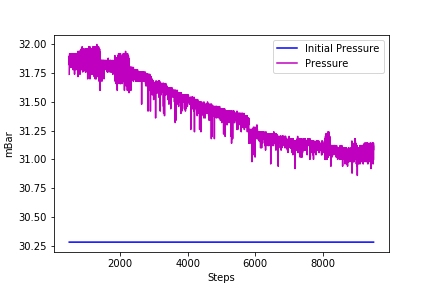}
    \caption{Pressure}\label{sn-result-0}
  \end{minipage} 
  \begin{minipage}[b]{0.5\linewidth}
    \includegraphics[width=0.85\textwidth]{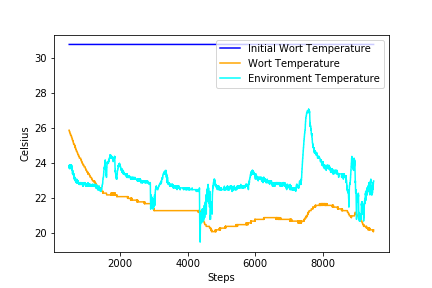}
    \caption{Temperatures}\label{sn-result-1}
  \end{minipage} 
  \begin{minipage}[b]{0.5\linewidth}
    \includegraphics[width=0.9\textwidth]{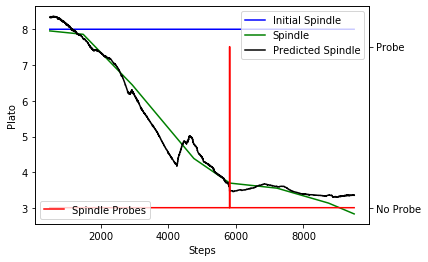}
    \caption{Spindle and predicted values}\label{sn-result-2} 
  \end{minipage}
  \hfill
  \begin{minipage}[b]{0.5\linewidth}
    \includegraphics[width=0.85\textwidth]{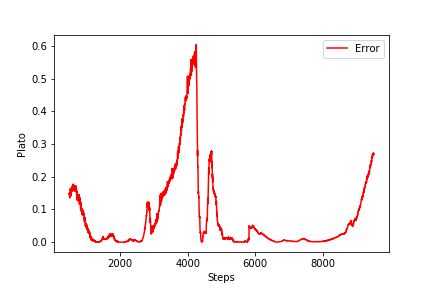}
    \caption{Error}\label{sn-result-3} 
  \end{minipage} 
\end{figure}

In Chapter~\ref{datagathering} and in Chapter~\ref{predictionfermentprocess}, the researchers have discussed the data gathering and data processing needed for training the LSTM network. In this chapter, the results after testing the trained LSTM network. Training was done with data from 24 fermentation processes. Data from seven fermentation processes were used for validating and testing. 
Figure~\ref{sn-result-0} and Figure~\ref{sn-result-1} show the pressure values and temperature values during a fermentation process. The respective sensor values used for testing have of course not been presented to the model during the training phase.

Just around Step 6000 a sudden drop of pressure occured which is a result of taking a spindle probe from the fermentation container. Figure~\ref{sn-result-2} shows the measured and extrapolated spindle values and the predicted spindle values from the neural network. At around Step 4000 predicted spindle values are increasing, an influence of the ambient and wort temperatures at exactly this time seems to be be responsible for that. 
Both, the actual spindle curve and the predicted spindle curve, are close to each other within an error of 0.6 Plato, see Figure~\ref{sn-result-3}. The error curve shows the least error values, which are calculated by taking the absolute values of the difference between the actual and predicted spindle curves. 

\begin{figure}[h]  
  \begin{minipage}[b]{0.5\linewidth}
    \includegraphics[width=0.85\textwidth]{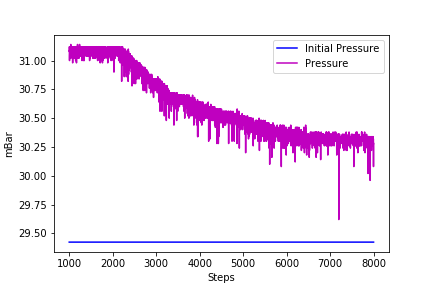}
    \caption{Pressure}\label{sn-result-4}
  \end{minipage} 
  \begin{minipage}[b]{0.5\linewidth}
    \includegraphics[width=0.85\textwidth]{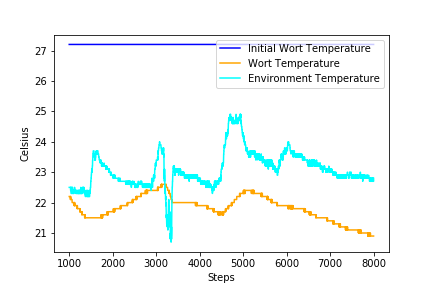}
    \caption{Temperatures}\label{sn-result-5}
  \end{minipage} 
  \begin{minipage}[b]{0.5\linewidth}
    \includegraphics[width=0.9\textwidth]{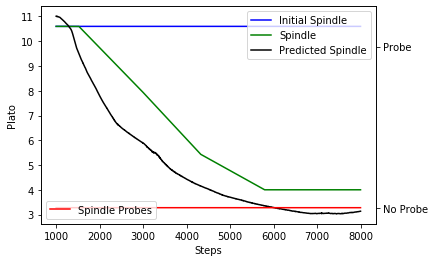}
    \caption{Spindle and predicted values}\label{sn-result-6} 
  \end{minipage}
  \hfill
  \begin{minipage}[b]{0.5\linewidth}
    \includegraphics[width=0.85\textwidth]{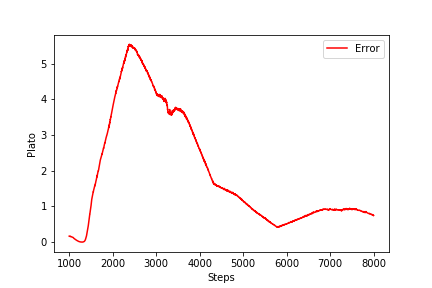}
    \caption{Error}\label{sn-result-7} 
  \end{minipage} 
\end{figure}

The following experiment is clearly a case where the LSTM network has failed the prediction.
In Figure~\ref{sn-result-4} and Figure~\ref{sn-result-5} the reader finds the pressure and temperature values. Only two spindle measurements were taken, one in the beginning and one at the end of the fermentation run (the graph does not show the complete curve, so the spindle events at beginning and end are not visible). The actual spindle curve was purely extrapolated by refractometer measurement values. Since there were no spindle measurements taken between the beginning and the end of the fermentation process, the pressure curve of Figure~\ref{sn-result-4} shows no sudden drops of pressure. The reader can see the actual and predicted spindle values in Figure~\ref{sn-result-6}. Figure~\ref{sn-result-7} shows the least absolute error curve, which exceeds 5 Plato between Step 2000 and Step 5000.

\begin{figure}[h]  
  \begin{minipage}[b]{0.5\linewidth}
    \includegraphics[width=0.85\textwidth]{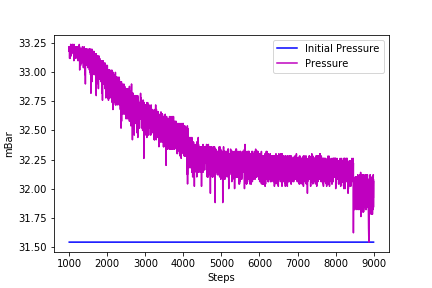}
    \caption{Pressure}\label{sn-result-8}
  \end{minipage} 
  \begin{minipage}[b]{0.5\linewidth}
    \includegraphics[width=0.85\textwidth]{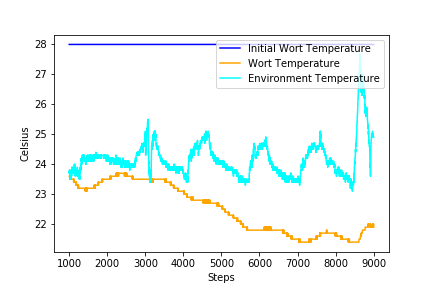}
    \caption{Temperatures}\label{sn-result-9}
  \end{minipage} 
  \begin{minipage}[b]{0.5\linewidth}
    \includegraphics[width=0.9\textwidth]{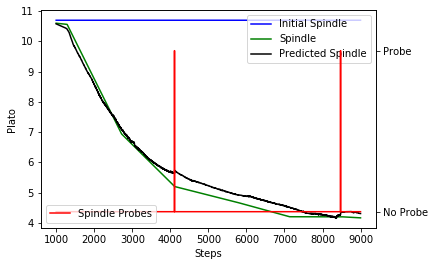}
    \caption{Spindle and predicted values}\label{sn-result-10} 
  \end{minipage}
  \hfill
  \begin{minipage}[b]{0.5\linewidth}
    \includegraphics[width=0.85\textwidth]{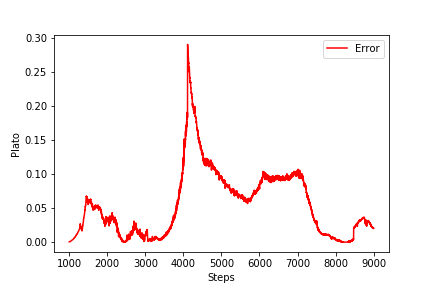}
    \caption{Error}\label{sn-result-11} 
  \end{minipage} 
\end{figure}

The data of the following experiment were used in the training process, so actual and predicted spindle values are expected to be close. Figure~\ref{sn-result-8} and Figure~\ref{sn-result-9} show the pressure and temperature values. At Step 4000 and Step 8000 the reader can see clearly the drops in the pressure curve, when the researchers have taken spindle probes from the fermentation container. Figure~\ref{sn-result-10} shows the actual and predicted spindle curves, which are very close to each other. This is no surprise, since the test data were used for training. The least absolute error in Figure~\ref{sn-result-11} is always below 0.3 Plato.

\section{Conclusion}\label{conclusion}

During training and validating the LSTM network, the researchers came to the conclusion that the initial values of the pressures are essential. The pressure sensor is actually measuring the pressure, but it can be used to weigh the wort at the start of the fermentation process. Wort having more weight often contains more fermentable sugar, and therefore more alcohol and carbon dioxide is produced. This results, that more weight is lost during fermentation. It must be noted the weight alone is not sufficient, since the wort density (indirectly sugar content) can vary in the beginning of the fermentation process, and is dependent on the previous brewing process. Because of this, the researchers added another initial value, which is the initial density, which the lab assistants measure right before the fermentation starts. To summarize, altogether the researchers have added two more features to the data frames, which are initial value of the pressure and the initial value of the spindle measurement. In Figure~\ref{sn-result-8} and Figure~\ref{sn-result-10} the reader finds both initial values as simple straight lines in blue.

\begin{figure}[h]%
\centering
\includegraphics[width=0.45\textwidth]{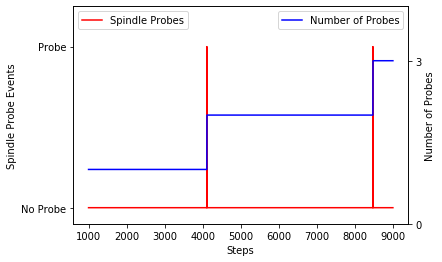}
\caption{Probe Events and Step Function}\label{sn-conclusion-0}
\end{figure}

The researchers also modified the spindle event feature in the data frames during the project. Each time a spindle probe has been taken (100ml of wort), content (and therefore weight) is removed. This can be seen e.g. in Figure~\ref{sn-result-8}, where the reader finds a drop of the pressure at this event. The pressure will not go back to the original level anymore. This means the researchers have to consider probe events from the past in the data frames. This can be done with replacing the probe event feature with data describing how many probes were taken. A step function is introduced with the number of probes containing information about the past and about the event itself. Figure~\ref{sn-conclusion-0} shows how an event curve is simply converted to a step function by summing up its events over the time.  

Until the end of the project, the researchers have executed 31 fermentation processes. Since the data of 7 processes are used for validation and testing, data from 24 processes were taken for training. The low number of processes is due to the fact, that one brewing process takes about four to six hours to execute and one fermentation process about 5-7 days. Often, the predictions work well, with errors below 0.6 Plato, but in other cases, we have errors above 5.0 Plato. The researchers deduct that the errors are due to the scarcity of data. More data might help to get better prediction. Another approach is to transfer a trained LSTM networks from other research projects, as proposed in \cite{bib28}, and retrain the network. This is due to the lack of similar general conditions between the projects difficult. 

In the following, the researchers suggest more for future work: 
Brewers keep the yeast during the fermentation process in a stable condition by controlling the wort's temperature with a generator cooling system. Since Figure~\ref{sn-result-1} and Figure~\ref{sn-result-2} show an impact of the fermentation temperature on the prediction values, a generator cooling system is recommended.
Currently, airlocks were used to control the pressure of the atmosphere inside the fermentation container. The researchers suggest to use a pressure controlled fermentation container, which can deliver a controlled atmosphere. Professional breweries often use these systems as well.
In the future, the authors would refrain from using the spindle, instead using a precise refractometer for wort density measurement. The spindle causes a drop in pressure, since 100ml wort is taken out the fermentation container. However, the refractometer just needs few drops of wort. Important is here to use an accurate refractometer. This measurement would make the above discussed feature "number of probes" inside the data frame obsolete.
There are a huge number of possibilities to vary the brew process. One variation can be to brew with different amounts of malt, which influences the wort density after the brewing process has finished. In this project, the researchers actually produced wort with initial densities of 8 Plato, 10 Plato and 12 Plato. Certainly, the LSTM network will not predict good results if the researchers experiment with different wort densities. This means more variations are needed for LSTM network training.  

Currently, only pressure sensors and temperature sensors are used to deduct the wort density. The main reason for the use of these sensors is their price. An extension of the sensor might be useful to improve the prediction, such as with electronic noses. In the articles \cite{bib29} and \cite{bib30} electronic noses are suggested to be used to monitor the alcohol content.

\section{Acknowledgement}\label{acknowledgement}

The \textit{Ministerium für Wirtschaft, Arbeit und Tourismus} in Baden-Württemberg, Germany financed this project within the program \textit{Initiative Wirtschaft 4.0}. Also, special thanks to the University of Applied Science Albstadt-Sigmaringen for providing facilities and basic equipment to enable this project.

\bibliographystyle{abbrv}                 
\nocite{*}
\bibliography{art-bibliography}

\begin{thebibliography}{10}

\bibitem{bib19}
Mcgraw-hill dictionary of scientific \& technical terms, 2003.

\bibitem{bib14}
P.~Attri, Y.~Sharma, K.~Takach, and F.~Shah.
\newblock Timeseries forecasting for weather prediction.
\newblock Keras Tutorial, 2020.
\newblock
  \url{https://keras.io/examples/timeseries/timeseries_weather_forecasting}.

\bibitem{bib10}
S.~A. Babichev, J.~Ries, and A.~I. Lvovsky.
\newblock Quantum scissors: teleportation of single-mode optical states by
  means of a nonlocal single photon, 2002.
\newblock Preprint at \url{https://arxiv.org/abs/quant-ph/0208066v1}.

\bibitem{bib11}
M.~Beneke, G.~Buchalla, and I.~Dunietz.
\newblock Mixing induced {CP} asymmetries in inclusive {B} decays.
\newblock {\em Phys. {L}ett.}, B393:132--142, 1997.

\bibitem{bib5}
M.~Broy.
\newblock Software engineering---from auxiliary to key technologies.
\newblock In M.~Broy and E.~Denert, editors, {\em Software Pioneers}, pages
  10--13. Springer, New {Y}ork, 1992.

\bibitem{bib1}
S.~L. Campbell and C.~W. Gear.
\newblock The index of general nonlinear {D}{A}{E}{S}.
\newblock {\em Numer. {M}ath.}, 72(2):173--196, 1995.

\bibitem{bib8}
S.~T. Chung and R.~L. Morris.
\newblock Isolation and characterization of plasmid deoxyribonucleic acid from
  streptomyces fradiae, 1978.
\newblock Paper presented at the 3rd international symposium on the genetics of
  industrial microorganisms, University of {W}isconsin, {M}adison, 4--9 June
  1978.

\bibitem{bib20}
C.~Cimander, M.~Carlsson, and C.-F. Mandenius.
\newblock Sensor fusion for on-line monitoring of yoghurt fermentation.
\newblock {\em Journal of Biotechnology 99 (2002) 237-248}, 2002.

\bibitem{bib28}
F.~Curreri, L.~Patane, and M.~G. Xibilia.
\newblock Rnn- and lstm-based soft sensors transferability for an industrial
  process.
\newblock {\em sensors}, 2021.

\bibitem{bib16}
H.~M. Esslinger, editor.
\newblock {\em Handbook of Brewing, Processes, Technology, Markets}.
\newblock Wiley-VCH Verlag GmbH \& Co. KGaA, Weinheim, 2009.

\bibitem{bib4}
K.~O. Geddes, S.~R. Czapor, and G.~Labahn.
\newblock {\em Algorithms for {C}omputer {A}lgebra}.
\newblock Kluwer, Boston, 1992.

\bibitem{bib29}
M.~Ghasemi-Varnamkhastiab, S.~Mohtasebi, M.~Rodriguez-Mendez, J.~Lozanoc,
  S.~Razavid, and H.Ahmadia.
\newblock Potential application of electronic nose technology in brewery.
\newblock {\em Trends in Food Science \& Technology}, 22:165--174, 2011.

\bibitem{bib3}
C.~Hamburger.
\newblock Quasimonotonicity, regularity and duality for nonlinear systems of
  partial differential equations.
\newblock {\em Ann. Mat. Pura. Appl.}, 169(2):321--354, 1995.

\bibitem{bib9}
Z.~Hao, A.~AghaKouchak, N.~Nakhjiri, and A.~Farahmand.
\newblock Global integrated drought monitoring and prediction system (gidmaps)
  data sets, 2014.
\newblock figshare \url{https://doi.org/10.6084/m9.figshare.853801}.

\bibitem{bib17}
K.-U. Heyse, editor.
\newblock {\em Praxishandbuch der Brauerei}.
\newblock Behr's Verlag.

\bibitem{bib12}
M.~Hoffmann, D.~Rembold, B.~Stauß, S.~Schwarzkopf, and M.~Jacobi.
\newblock Konzeption und aufbau einer trainingsinfrastruktur für virtuelle
  sensoren.
\newblock {\em Anwendungen und Konzepte der Wirtschaftsinformatik}, 15, 2022.
\newblock Print at
  \url{https://ojs-hslu.ch/ojs3211/index.php/akwi/issue/view/15}.

\bibitem{bib31}
W.~Ke, D.~Huang, F.~Yang, and Y.~Jiang.
\newblock Soft sensor development and applications based on lstm in deep neural
  networks.
\newblock In {\em 2017 IEEE Symposium Series on Computational Intelligence
  (SSCI)}, Honolulu, 2017. IEEE.

\bibitem{bib24}
C.~F. Lui, Y.~Liu, and M.~Xie.
\newblock A supervised bidirectional long short-term memory network for
  data-driven dynamic soft sensor modeling.
\newblock {\em IEEE TRANSACTIONS ON INSTRUMENTATION AND MEASUREMENT, VOL. 7},
  2022.

\bibitem{bib21}
C.~Murugan.
\newblock Soft sensors for biomass monitoring during low cost cellulase
  production, 2021.

\bibitem{bib23}
N.~A. Onate, C.~A. Garcia, and M.~A. Beltran.
\newblock A fermenter model based on neural networks experimentally validated.
\newblock In {\em 2017 IEEE International Conference on Industrial Technology
  (ICIT)}, Toronto, 2017. IEEE.

\bibitem{bib22}
H.~Pan, T.~Su, X.~Huang, and Z.~Wang.
\newblock Lstm-based soft sensor design for oxygen content of flue gas in
  coal-fired power plant, 2020.

\bibitem{bib27}
M.~Petrova, P.~Koprinkova, and T.~Patarinska.
\newblock Neural network modelling of fermentation processes. microorganisms
  cultivation model.
\newblock {\em Bioprocess Engineering 16}, 1997.

\bibitem{bib30}
J.~P. Santos, J.~Lozano, and M.~Aleixandre.
\newblock Electronic noses applications in beer technology, 2017.

\bibitem{bib6}
R.~S. Seymour, editor.
\newblock {\em Conductive {P}olymers}.
\newblock Plenum, New {Y}ork, 1981.

\bibitem{bib25}
A.~Sipos, A.~Florea, M.~Arsin, and U.~Fiore.
\newblock Using neural networks to obtain indirect information about the state
  variables in an alcoholic fermentation process.
\newblock {\em processes}, 2020.

\bibitem{bib2}
M.~K. Slifka and J.~L. Whitton.
\newblock Clinical implications of dysregulated cytokine production.
\newblock {\em J. {M}ol. {M}ed.}, 78:74--80, 2000.

\bibitem{bib7}
S.~E. Smith.
\newblock Neuromuscular blocking drugs in man.
\newblock In E.~Zaimis, editor, {\em Neuromuscular junction. {H}andbook of
  experimental pharmacology}, volume~42, pages 593--660, Heidelberg, 1976.
  Springer.

\bibitem{bib13}
B.~Stahl.
\newblock deep{SIP}: deep learning of {S}upernova {I}a {P}arameters.
\newblock Astrophysics {S}ource {C}ode {L}ibrary, Jun 2020.

\bibitem{bib15}
tensorflow.
\newblock Time series forecasting.
\newblock Tensorflow Core.
\newblock
  \url{https://www.tensorflow.org/tutorials/structured_data/time_series}.

\bibitem{bib18}
S.~Yu, G.~Montague, and E.~Martin.
\newblock Data fusion for enhanced fermentation process tracking.
\newblock In {\em 9th International Symposium on Dynamics and Control of
  Process Systems (DYCOPS 2010)}, Leuven, 2010. IFAC.

\bibitem{bib26}
X.~Zhu1, K.~U. Rehman, W.~Bo, M.~Shahzad1, and A.~Hassan.
\newblock Data‑driven soft sensor model based on deep learning for quality
  prediction of industrial processes.
\newblock {\em SN Computer Science}, 2021.

\end{thebibliography}

\end{document}